\documentclass{esannV2}
\usepackage{graphicx} [dvips]
\usepackage[latin1]{inputenc}
\usepackage{amssymb,amsmath,array, mathabx, url, amsthm, algorithm, algpseudocode}
\usepackage[numbers]{natbib}
\theoremstyle{definition}
\newtheorem{definition}{Definition}

\begin{document}
%style file for ESANN manuscripts
\title{Closed-loop multi-step planning with innate physics knowledge}

%***********************************************************************
% AUTHORS INFORMATION AREA
%***********************************************************************
\author{Giulia Lafratta$^1$, Bernd Porr$^1$, Christopher Chandler$^2$ and Alice Miller$^2$
%
% Optional short acknowledgment: remove next line if non-needed
\thanks{This work was supported by a grant from the UKRI Engineering and Physical Sciences Research Council Doctoral Training Partnership award [EP/T517896/1-312561-05]; the UKRI Strategic Priorities Fund to the UKRI Research Node on Trustworthy Autonomous Systems Governance and Regulation [EP/V026607/1, 2020-2024]; and the UKRI Centre for Doctoral Training in Socially Intelligent Artificial Agents [EP/S02266X/1].}
%
% DO NOT MODIFY THE FOLLOWING '\vspace' ARGUMENT
\vspace{.3cm}\\
%
% Addresses and institutions (remove "1- " in case of a single institution)
1- University of Glasgow - School of Engineering\\
James Watt South Building, Glasgow G12 8QQ - United Kingdom
%
% Remove the next three lines in case of a single institution
\vspace{.1cm}\\
2- University of Glasgow - School of Computing Science\\
18 Lilybank Gardens, Glasgow G12 8RZ - United Kingdom\\
}
%***********************************************************************
% END OF AUTHORS INFORMATION AREA
%***********************************************************************

\maketitle

\begin{abstract}
We present a hierarchical framework to solve robot planning as an input control problem. At the lowest level are temporary closed control loops, (``tasks''), each representing a behaviour, contingent on a specific sensory input and therefore temporary. At the highest level, a supervising ``Configurator'' directs task creation and termination. Here resides ``core'' knowledge as a physics engine, where sequences of tasks can be simulated. The Configurator encodes and interprets simulation results, based on which it can choose a sequence of tasks as a plan.
We implement this framework on a real robot and test it in an overtaking scenario as proof-of-concept.
\end{abstract}

\section{Introduction}
Living organisms interact with their surroundings through sensory inputs in a closed-loop fashion \citep{Maturana1980}. 
To achieve basic closed-loop navigation in a robot it is sufficient to directly connect a robot's sensors to its motor effectors, and the specific excitatory or inhibitory connections determine the control strategy~\citep{Braitenberg1986}. These sensor-effector connections represent a single closed-loop controller, which produces stereotyped behaviour in response to an immediate stimulus. 

The state-of-the-art for autonomous navigation falls into the categories of closed-loop \textit{output} control (e.g. Reinforcement Learning, RL), or trajectory planning (see~\cite{Karur2021}). In closed-loop output control, an agent uses environmental feedback to learn to associate input states with actions, aiming to maximise the cumulative reward. Feedback is provided at the end of each training episode, which makes a RL model inherently slow to train. Trajectory planners represent, on the other hand, open-loop controllers. Environment feedback may be provided on trajectory \textit{following}. As an embodied agent is oblivious to its trajectory, this requires an external observed (e.g. a camera on the ceiling). 

Closed-loop behaviours (CLB) have some advantages over the state-of-the-art. First, the behaviour displayed by a robot depends on hard-wired connections; thus, responses to inputs do not need to be learned from scratch through lengthy training. 
Secondly, to implement CLB, a robot only needs information about the location of an object relative to itself to successfully perform a control manoeuvre. On the other hand, as a single control loop only receives information about an object at a time,
it is not, by itself, suitable for the formulation of complex navigation plans. \citet{Spelke2007} introduce ``core knowledge'' as the seemingly innate understanding of basic properties of one's environment such as physics and causality observed in newborn animals. In this work, we propose a framework where CLB such as described in~\citep{Braitenberg1986} are combined with core knowledge to achieve multi-step ahead planning in an embodied, situated agent.

\section{Methods} \label{theory}
\subsection{Preliminaries}
\subsubsection{Tasks as closed-loop controllers}
\begin{figure}[htb!]
\centering
\includegraphics[width=0.5\textwidth]{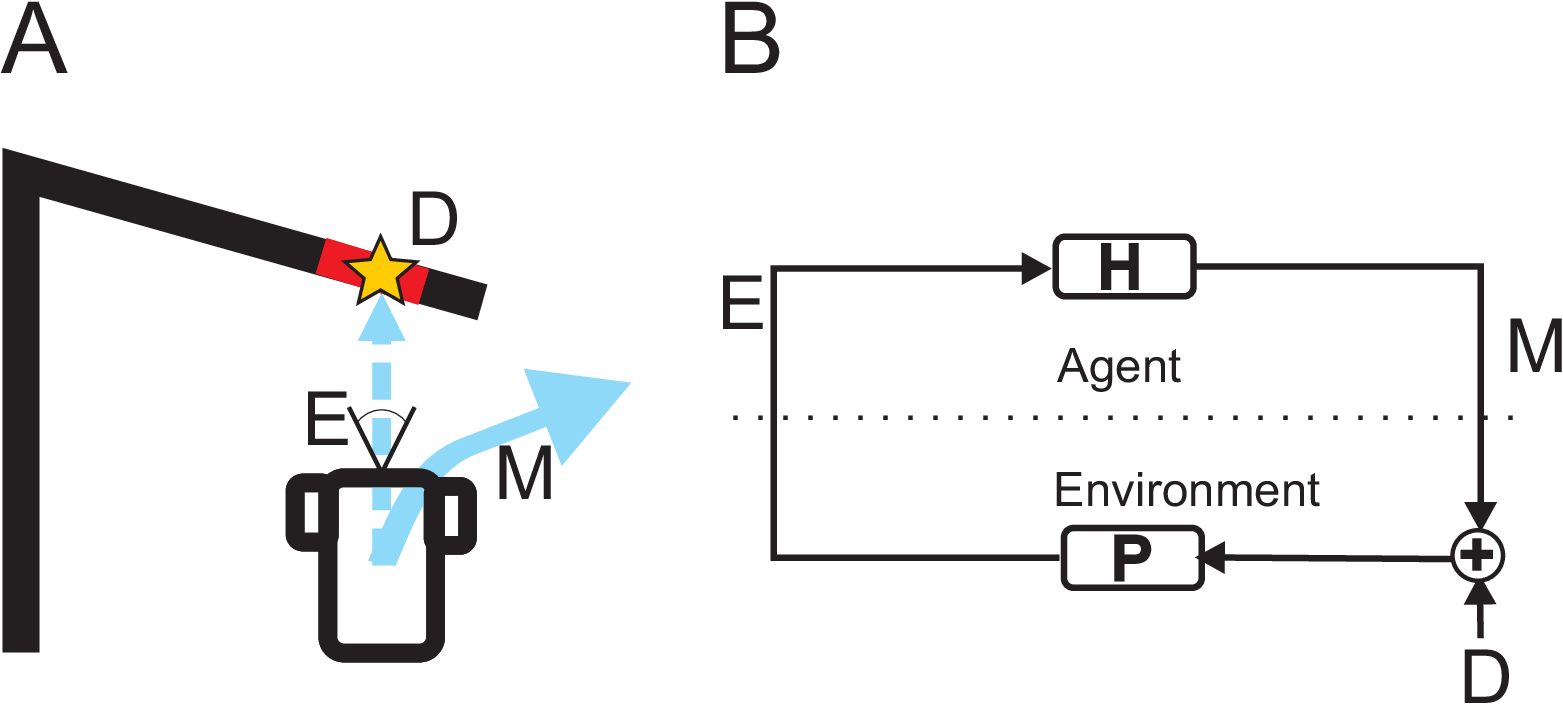}
    \caption{A: closed-loop obstacle avoidance. B: a closed-loop controller.}
    \label{fig:task}
\end{figure}

Fig.~\ref{fig:task} provides a visual representation of a task. As shown in Fig.~\ref{fig:task}B, a task is a closed-loop behaviour contingent on a disturbance $D$ which enters the environment $P$ and generates an error signal $E$ via the robot's sensors. As long as signal $E$ is nonzero, the agent $H$ produces a continuous motor output $M$ aimed at counteracting the disturbance (see Fig.~\ref{fig:task}A). Otherwise, the control behaviour is no longer necessary and it terminates. 

\subsubsection{The Configurator}\label{conf_informal}
 \label{conf_form}
\begin{figure}[htb!]
    \centering
    \includegraphics[width=\textwidth]{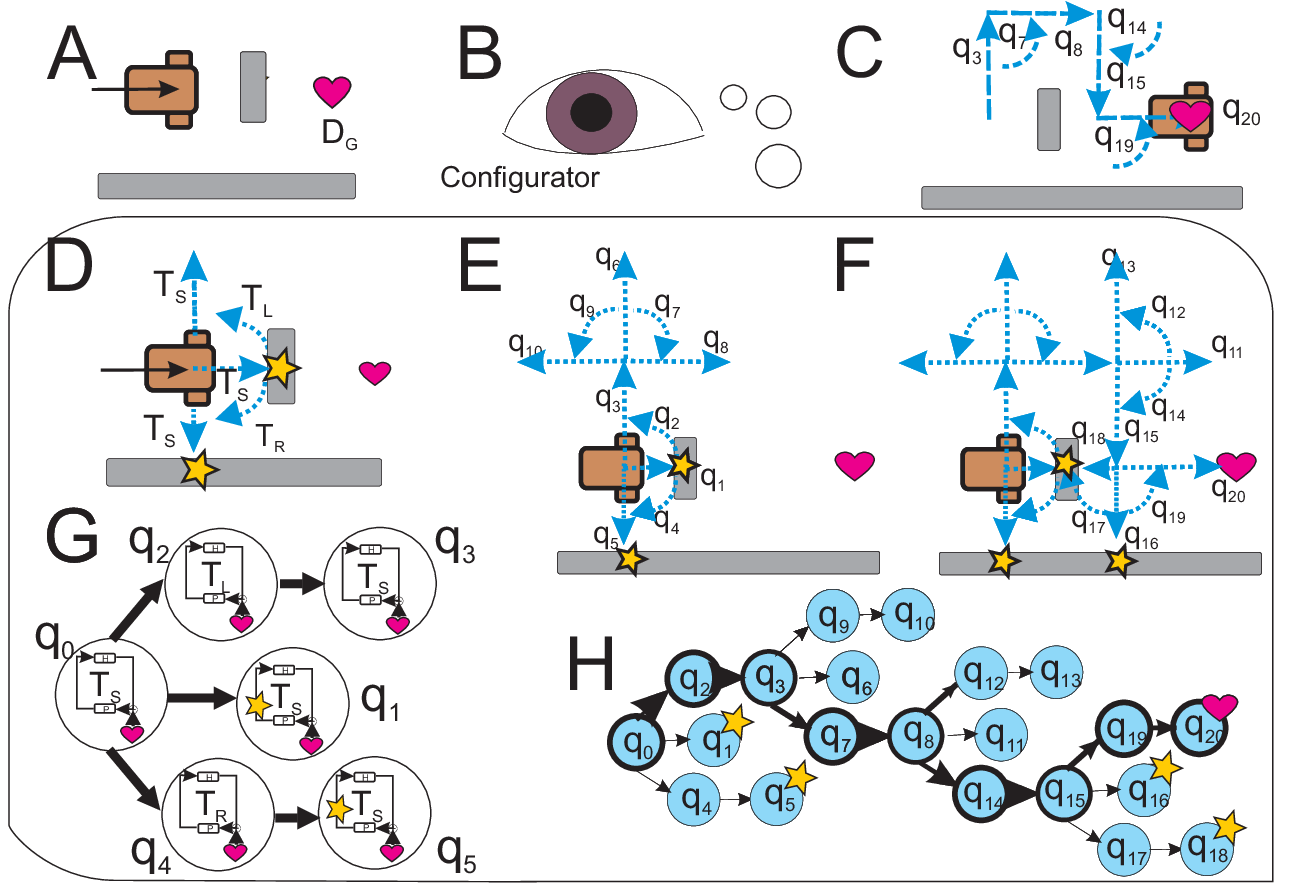}
    \caption{ 
    Example of the multi-step-ahead planning procedure carried out by the Configurator. A: a scenario requiring multi-step planning, B: Configurator, C: the plan, D: expansion of the start state, E: expansion of bext next state, F: full state-space expansion, G: state diagram of panel E, H: full state diagram.}   
    \label{fig:react_loop}
\end{figure}

Fig.~\ref{fig:react_loop} illustrates the role of the Configurator in the planning process. Fig.~\ref{fig:react_loop}A presents a scenario requiring multi-step planning: a robot is driving straight (task $T_S$) aimed at reaching the target $D_G$ in an environment where obstacles are present in front of it and to its right. The Configurator (Fig.~\ref{fig:react_loop}B) thus initiates planning. First, the environment has to be explored in order to know what tasks are feasible to perform to reach the goal. As denoted by the thought balloon, this exploration is not actual but simulated using core knowledge. Fig.~\ref{fig:react_loop}D presents the first stage of the exploration process. From the present task, the robot has the option to proceed straight (task $T_S$), turn left (task $T_L$) and then drive straight, or turn right (task $T_R$) and then go straight. The first and second options result in collisions (marked as stars) with the obstacle in front, while the third can be executed safely. These task sequences are represented in Fig.~\ref{fig:react_loop}E: the control strategies represented by each task ($T_S, T_L, T_S$) are inscribed in the corresponding control loop. To motivate the existence of a CLB, the Configurator injects the goal $D_G$ as a disturbance in each task. Tasks whose execution is interrupted by a collision are marked with stars. Tasks and disturbances associated with them are collected in structures $q$ containing tasks and their disturbances, where the subscript indices indicate the order of task creation. As shown in figure~\ref{fig:react_loop}F, the next stage of exploration starts from the task and disturbances in structure $q_3$ in a modular fashion following the steps in Fig.~\ref{fig:react_loop}E. Fig.~\ref{fig:react_loop}G depicts the full extent of the state-space expansion. Note that after one bout of modular (simulated) exploration, even if no collisions are encountered (see states $q_6, q_8, q_10$, the exploration starts from the state which brings the robot closest to the goal (in this case $q_8$). Fig.~\ref{fig:react_loop}H depicts the collection of the explored task sequences in form of a cognitive map. This structure is searchable and a plan (bold arrows) can be extracted. Each task is queued for execution on the real robot, as shown in Fig.~\ref{fig:react_loop}C.
\subsubsection{Formalism}
%\begin{definition}
   % The state space is $Q:\mathbf{T} \times 2^{\mathbf{D}}$, where $\mathbf{T}$ is the task space and $2^{\mathbf{D}}$ is the power set of $\mathbf{D}$, a set of disturbances. 
%\end{definition}
\begin{definition}
    A state is a tuple $q=(T, D)$, where $T$ is a task and $D$ is a finite, possibly empty, set of disturbances.
\end{definition}
In set $D$ we use $D_I$ to indicate the disturbance to which the task is contingent, and $D_N$ to define a disturbance which interrupts the task execution.
\begin{definition}
     Set $\mathbf{T}$ is the task space such that $T \in \mathbf{T}$. Set $\mathbf{D}$ is the set of all disturbances such that $D \in \mathbf{D}$.
\end{definition}
\begin{definition}\label{ts}
    A cognitive map is a tuple $(Q, \hookrightarrow)$, where $Q:\mathbf{T} \times 2^{\mathbf{D}}$ is a set of states and $\hookrightarrow:Q \times Q$ is a set of transition relations.
\end{definition}

\begin{definition}{
The Configurator is a tuple
%\begin{equation*}
    $(\mathcal{G},D_G, \Psi, R)$
%    \label{conf_eq}
%\end{equation*}
where $\mathcal{G}=(Q, \hookrightarrow)$ is a cognitive map representing the searchable state-space, $D_G$ is the system's overarching goal (a disturbance which the \textsl{plan} aims to counteract), function \(\Psi:\hookrightarrow \rightarrow [1,0]\) is a guard which assigns a supervisory control pattern~\citep{Ramadge1984} to each transition, 
\(R: \hookrightarrow \rightarrow \mathbf{D}\) is a reset through which the Configurator injects disturbances from disturbance set $\mathbf{D}$ into states in a top-down fashion. }
\end{definition}

\subsection{Implementation}
\subsubsection{Core knowledge}
The physics engine Box2D\footnote{\url{https://box2d.org/}} represents the core knowledge using which sequences of tasks can be simulated. 
Objects in the Box2D simulation are constructed from ego-centric coordinates captured by the robot's LiDAR sensor. The raw point cloud is resampled so that only the points in the way of the present task are represented.
%The extreme coordinates in this subset determine the object's dimensions. 
A model of the robot is also present in the environment. At the beginning of the simulation, the robot is located at the origin of the plane 
\subsubsection{Reset}
In this work, we define reset 
\begin{equation*}
R(q_1, q_2):D'_I=
 D_G \ \text{if} \ D_N= \emptyset
\end{equation*}
where $D'_I$ is the disturbance to be injected in the task in state $q_2$, and $D_N$ is new disturbance which interrupted the task in state $q_1$. In other words, unless the task in $q_1$ is interrupted, the next task in $q_2$ will be contingent on the goal $D_G$. As we seek to plan to \textit{avoid} collisions, transitions from tasks where set $D_N\neq \emptyset$ are not permitted. The reset for this case is not defined.
\subsubsection{Construction of cognitive map $\mathcal{G}$}
As mentioned in Section~\ref{conf_informal}, we use a best-first approach to assign priority of expansion to states. As in~\citet{Hart1968}, we define a cost function $\phi: Q \rightarrow \mathbb{R}$ such that
\begin{equation*}
    \phi(q) = \gamma(q)+\chi(q) \label{ev_function} 
\end{equation*}
where $q$ is a state in cognitive map $\mathcal{G}$, $\gamma(q)$ is a past cost function and $\chi(q)$ is a heuristic future cost function. We calculate $\gamma(q)$ as the normalised distance from the start of the task to interrupting disturbance $D_N$, if present, and $\chi (q)$ as the normalised negative distance from the end of the task to goal $D_G$. Algorithm~\ref{e} provides pseudocode for the construction of cognitive map $\mathcal{G}$. 

\begin{algorithm}
    \caption{Construction of the cognitive map $\mathcal{G}$} 
    \begin{algorithmic}
        \State inputs: $\mathbf{D}_G$,
        \State initialise $Q={q_0}$, priority queue $PQ=\{q_{0}\}$, state to expand $q_{E}=q_{0}$, 
        \Repeat
        \State remove \(q_E\) from \(PQ\)
        \State simulate all frontier states $q_F$ out of $q_E$ depth-first until $q_F=(T_S, \cdot)$ or $D_N \neq \emptyset$%,\\
        %\hfill every state initialised with $D'_I = D_G$
        \State add $q_{F}$ to $PQ$
        \State set \(q_{E} = \min_{q \in PQ}\phi(q)\)
        \Until{no obstacle-free transitions exist out of $q_{E}$ or $D_{G}$ has been reached}\\
    \Return \(\mathcal{G}\)
    \end{algorithmic}\label{e}
\end{algorithm}
%where $PQ$ is a priority queue, $q_E$ is the state whose out-transitions are being explored, and $q_F$ is a frontier state.
\subsubsection{Guard $\Psi$}
Guard $\Psi$ is used to extract a plan $\rho$ from cognitive map $\mathcal{G}$. Formally, the plan is
\begin{equation}
    \rho = \{q_0, q_1...q_n  \ | \ (q_i, q_{i+1}) \in \hookrightarrow \forall i \in \mathbb{N} \ , \  q_n= \min_{q \in Q} \phi(q)\}
\end{equation}
or a set of adjacent states where the last state has the least cost $\phi$ in the set of states $Q$. This results in the assignment of $\psi(q)=1$ if $q\in \rho$ and zero otherwise.
\section{Results}
\begin{figure}
    \centering
    \includegraphics[width=.75\textwidth]{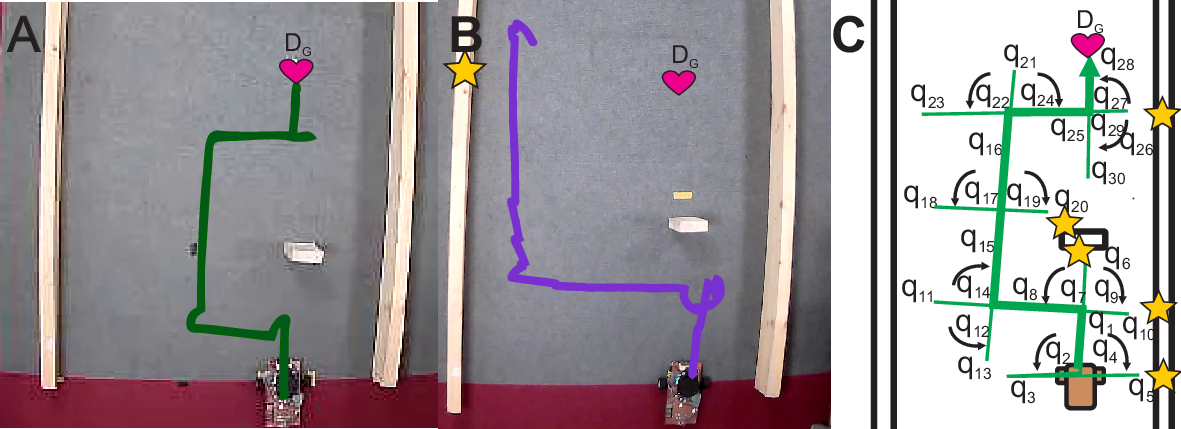}
    \caption{  Initial frame and tracking of the behaviour exhibited by the robot. Targets are depicted as hearts, collisions as stars. A: multiple-loop planning, B: one-loop-ahead disturbance detection, C: simulation trace and state labels for planning condition, executed plan highlighted in bold.  
    } \label{fig:tar_all}
\end{figure}

We test the decision-making process in an overtaking scenario, where a target location $D_G$ located 1 meter in front of the robot must be reached in presence of an obstacle. We contrast our new approach against a reactive control strategy which uses a single closed-loop controller. In the planning condition, the robot was able to formulate a plan to reach the target in all 10 runs (example run depicted in Fig.~\ref{fig:tar_all}A). The mean time for state-space exploration and planning was 0.064 $\pm$ 0.009 seconds for a state-space comprised of 32.9 $\pm$ 1.758 states (see Fig.~\ref{fig:tar_all}C for an example simulation trace with state labels). During execution, the robot brushed the obstacle twice in the planning condition, both times in one particular scenario due to execution error accumulation. The robot never reached the target in the reactive condition (see Fig.~\ref{fig:tar_all}B); moreover, it collided 8 times with obstacles over the ten runs. Differences in number of collisions approached, but did not reach, statistical significance (paired T=-1.747, p=0.111), possibly due to the small sample size. 

\section{Conclusion}
We have developed a framework to achieve multi-step ahead, real-time planning over CLB thanks to core knowledge as a physics simulation. This work showcases an innovative use of a physical simulation as a tool for an agent to actively reason over its environment. This allows for constructing a cognitive map without need for physical exploration, where the environment is discretised in terms of CLB afforded by objects in the environment.

\begin{footnotesize}
\bibliographystyle{unsrtnat} %unsrt
\bibliography{export}
\end{footnotesize}

\end{document}